# A Fine Evaluation Method for Cube Copying Test for Early Detection of Alzheimer's Disease

Xinyu Jiang[1], Cuiyun Gao[1*], Wenda Huang[1], Yiyang Jiang[1], Binwen Luo[1], Yuxin Jiang[1]
Mengting Wang[1], Haoran Wen[1], Yang Zhao[1], Xuemei Chen[2], Songqun Huang[3]

[1]General Purposed Auto-Test & AI Tech. Lab, Anhui Jianzhu University, Hefei, 230601, China

[2]Hefei Smart Healthcare Center & Apartment for Elderly, Hefei Industrial Investment Co., LTD
Hefei, 230011, China

[3]Department of Cardiovasology Changhai Hospital, Second Military Medical University,
Shanghai, 200433, China

**Abstract:**

Background: Impairment of visual spatial cognitive function is the most common early clinical manifestation of Alzheimer's Disease (AD). When the Montreal Cognitive Assessment (MoCA) uses the "0/1" binary method ("pass/fail") to evaluate the visual spatial cognitive ability represented by the Cube Copying Test(CCT), the elder with less formal education generally score 0 point, resulting in serious bias in the evaluation results. Therefore, this study proposes a fine evaluation method for CCT based on dynamic handwriting feature extraction of DH-SCSM-BLA.

method : The Cogni-CareV3.0 software independently developed by our team was used to collect dynamic handwriting data of CCT. Then, the spatial and motion features of segmented dynamic handwriting were extracted, and feature matrix with unequal dimensions were normalized. Finally, a bidirectional long short-term memory network model combined with attention mechanism (BiLSTM-Attention) was adopted for classification.

Result: The experimental results showed that: The proposed method has significant superiority compared to similar studies, with a classification accuracy of 86.69%. The distribution of cube drawing ability scores has significant regularity for three aspects such as MCI patients and healthy control group, age, and levels of education. It was also found that score for each cognitive task including cube drawing ability score is negatively correlated with age. Score for each cognitive task including cube drawing ability score, but positively correlated with levels of education significantly.

Conclusion: This study provides a relatively objective and comprehensive evaluation method for early screening and personalized intervention of visual spatial cognitive impairment.

# 1. Introduction

## 1.1 Background

Alzheimer's Disease (AD) is a neurodegenerative disorder characterized by insidious onset and progressive development, which affects memory, thinking, and the ability to perform daily activities[1],[2]. It is estimated that 15.07 million people aged

---

\* Corresponding author.
E-mail addresses: jiangxinyu@stu.ahjzu.edu.cn (X. Jiang), gaocuiyun@ahjzu.edu.cn (C. Gao)

60 years or older in China have dementia: 9.83 million with Alzheimer's disease[3]. The number of people with dementia is projected to increase from 57.4 million cases globally in 2019 to 152.8 million cases in 2050[4]. Moreover, the low consultation rate among AD patients is primarily due to a lack of attention in the early stages of the disease. Consequently, many patients do not seek medical care until the condition has progressed to later phases, often missing the optimal window for intervention. Therefore, the management of AD must be guided by the principle of early detection and timely intervention. The early stage of the condition, known as Mild Cognitive Impairment (MCI), represents a transitional stage between being cognitively unimpaired and dementia, so consensus has been reached to focus primary interventions in this population to prevent dementia[3]. One of the most prominent early signs in MCI is the decline in cognitive function. Cognitive impairment affects skills such as communication, comprehension, or memory, handwriting is one of the daily activities affected by these kinds of impairments, and its anomalies are already used as diagnosis sign. Nowadays, many studies have been conducted to investigate how cognitive impairments affect handwriting[5]. Research has found that motor planning and cognitive ability appear to be more closely related to handwriting[6]. When Alzheimer's disease was first diagnosed in 1970, researchers had already found a positive correlation was observed between the severity of the dementia and handwriting measures[7]. Further research has revealed that the pathological mechanisms of dysgraphia are closely associated with impaired visuospatial cognitive ability. Cube copying measures visuospatial cognitive ability, which is often impaired in Alzheimer's disease[8].

To systematically quantify cognitive impairments, clinicians have developed multidimensional assessment tools including the Mini-Mental State Examination (MMSE), Montreal Cognitive Assessment (MoCA), and Addenbrooke's Cognitive Examination-III (ACE-III). These tests, designed primarily for assessing AD and other dementia-causing conditions, focus predominantly on cognitive domains affected in these disorders: episodic memory, spatiotemporal orientation, language, and higher visual processing[9]-[12]. Recent years have witnessed growing emphasis on handwriting analysis-based cognitive evaluation methods. Studies demonstrate significant differences in handwriting characteristics, speed, rhythm, and stability between Parkinson's disease (PD) patients, AD patients, and healthy individuals[13],[14]. Therefore, this study primarily investigates the impact of handwriting feature analysis methods in the CCT on cognitive assessment, utilizing multiple AD-related cognitive scales.

## 1.2 Related Work

This section focuses on the core objective of assessing visuospatial cognitive ability. It begins by surveying research on dynamic handwriting signals, analyzing the current status of studies from three dimensions: evaluation methods for visuospatial cognitive ability, feature extraction and classification methods based on static images, and feature extraction and classification methods based on dynamic trajectories. The aim is to achieve an assessment of visuospatial cognitive ability using dynamic handwriting signals.

## 1.2.1 Methods for assessing CCT based visuospatial cognitive ability

Visuospatial cognitive ability, a core dimension of cognitive function, encompasses object position perception, spatial relationship judgment, and visual-motor coordination. It serves as an early biomarker for neurodegenerative disorders such as AD, Lewy body dementia (DLB), and PD, holding significant clinical diagnostic value. Salimi et al.[15] demonstrated that visuospatial assessment can effectively enhance diagnostic accuracy for Alzheimer's disease, particularly during early disease stages, serving as a crucial supplementary tool for cognitive evaluation. CCT is a classic visuospatial cognitive ability assessment task and possesses unique diagnostic value. The CCT can also be viewed as a drawing process, which requires the integration of multiple cognitive resources to draw three-dimensional spatial figures, primarily involving visuospatial processing, eye-hand coordination, and other higher-order cognitive ability.

This process involves integrating multiple cognitive resources to construct three-dimensional spatial figures, primarily engaging visuospatial processing and hand-eye coordination—key advanced cognitive ability[16]. Shimada et al.[17] found that cube replication tasks are sensitive and effective neuropsychological tests, capable of revealing early visuospatial dysfunction and constructive apraxia in AD. Furthermore, the CCT has been incorporated into multiple standardized assessment scales. For instance, Sebastian Corral et al.[9] demonstrated that the MOCA scale quantifies visuospatial cognitive ability through a clock drawing task (3 scores) and a 3D cube copy task (1 scores). Andrew et al.[18] compared the diagnostic accuracy of ACE, ACE-R, and MMSE scales, finding that ACE-R evaluates visuospatial cognitive ability through cube replication with slightly better accuracy than MMSE. Additionally, Kazuya et al.[19] utilized subtests from the Visual Perception Function Scale (VOSP) – including cube/pentagon replication and clock drawing tasks – to assess visual perception abilities and investigate visual impairment characteristics in PD patients. Juan [20] employed the Rey-Osterrieth Complex Figure Test (ROCF) to assign quantitative scores based on geometric similarity between drawings and models, thereby evaluating participants' visual memory and spatial organization abilities. However, traditional assessment methods fail to capture dynamic cognitive processes. With the rapid development of digital technologies, dynamic handwriting analysis has emerged as a novel technique for visuospatial evaluation. This approach has also been applied to CCT, where motion handwriting data collected during CCT completion can quantify spatial perception, directional judgment, and motor coordination ability, thereby extracting quantifiable features that effectively characterize visuospatial cognitive ability.

## 1.2.2 Feature extraction methods based on static images

Feature extraction from static images aims to reduce the dimensionality of raw image data while preserving the most informative and discriminative features[21]. For cube copying images, this process involves extracting critical information including line structures, geometric shapes, and bending angles from the collected images to support subsequent research. For instance, Bernardo et al.[22] applied Convolutional Neural Network (CNN) model as a feature extractor and a support vector machine (SVM) as a classifier to analyze differences between handmade drawings of PD patients and healthy subjects. Pakize et al.[23] proposed using 1D features from tasks in the public dataset DARWIN to generate 2D RGB image features, which were yielded into the novel CNN model for early AD diagnosis analysis. Additionally, Wahiba Ismaiel et al.[24] extracted edge features from original images using Canny edge detection and enhanced the recognition of static Arabic sign language gestures via ResNet50 and the Agile Convolutional Neural Network (ASLR_CNN).

While static images are widely used, they cannot adequately capture temporal dynamics, limiting their effectiveness. This has positioned dynamic trajectory-based feature extraction as a key advancement, especially in fields like neurological disease assessment and behavior analysis, where the analysis of motion is crucial.

### 1.2.3 Feature extraction methods based on dynamic handwriting

Dynamic trajectories provide richer temporal information than static images, enabling more comprehensive analysis. Researchers extract relevant features from dynamic handwriting signals to provide crucial evidence for disease diagnosis and cognitive assessment. For instance, Nardone et al.[25] conducted independent analysis by integrating handwriting signals from multiple handwriting tasks, preserving fine motor information through stroke-by-stroke analysis to detect cognitive decline. Valla et al.[26] achieved efficient classification of PD patients by extracting kinematic and spatiotemporal features from drawing trajectories in the Archimedes spiral drawing test. Rios-Urrego et al.[27] systematically compared the modeling effects of three types of features—kinematic, geometric, and nonlinear dynamic features—on Parkinson's writing disorders, achieving a maximum classification accuracy of 93.1%.

In dynamic handwriting research, deep learning models are predominantly integrated. Traditional machine learning methods primarily employ classifiers such as K-nearest neighbor (KNN), decision trees (DT), and random forests (RF) [26]-[29]. With the advancement of deep learning technologies, classification and prediction methods utilizing deep models—including CNN, bidirectional gated recurrent units (BiGRU), and long short-term memory networks (LSTM)—have emerged as key research focuses in this field [30]-[35]. For example, Ma et al.[28] proposes a novel spatial temporal spectral fusion neural network (STSNet), the model can efficiently fuse complementary information from static handwriting images and dynamic multi-sensory fusion signals for the fine-grained assessment of tremor severity in PD. Cilia et al.[29] constructed a novel online handwriting dataset to analyze combinations of handwriting features for AD diagnosis. They utilized multiple machine learning algorithms, including DT, RF, KNN, and Logistic Regression (LR), for AD classification. Wang et al.[30] demonstrated that 3D CNN models outperformed 1D, 2D, and 3D-CNN models in capturing spatiotemporal dynamics, providing new insights for designing deep learning architectures for handwriting classification. Diaz et al.[31] developed a CNN-BiGRU hybrid model to evaluate the potential of dynamic handwriting analysis in PD. Ma et al.[32] combined a novel Transformer-based deep learning model with Archimedean spiral drawing tasks performed by patients and healthy subjects, effectively assessing its potential for diagnosing tremor symptoms. Vidhya R.[33] proposed integrating the feature extraction capabilities of ResNet with the temporal modeling advantages of LSTM, revealing the model's exceptional performance in identifying early AD biomarkers.

### 1.2.4 Relative Research Foundations of Our Team

Since 2021, our team has been conducting research on early detection and intervention for Alzheimer's disease. Based on comprehensive study of scales such as the MMSE and MOCA, we have developed two cognitive assessment apps [36] and a web-based testing platform. All three software have the function of remote cognitive function testing and automatic uploading of test data to cloud server. Meanwhile, for intervention studies, our team developed a brain health exercise software and proposed a fuzzy progress evaluation method [37], while also designing training software for scenario-based object retrieval based on VR. The software developed by our team, named Cogni-Care (V1.0, V2.0, and V3.0), including a total of 11 tasks assessing short-term memory, executive function, naming, visuospatial cognition, calculation, attention, language ability, orientation, abstraction, and delayed memory. The visuospatial cognition task includes clock drawing test and cube copying test. The intelligent speech recognition for language assessment was implemented based on the Iflytek Automatic Speech Recognition (ASR) Package, and the assessment of the CCT task is conducted using an intelligent algorithm that was developed independently by our team.

During data collection, we found that most of participants aged over 70 had less formal education, who lacked the concept of solid geometry or three-dimensional (3D) and also lacked training in cube copying. As a result, the cube images they draw generally lacked 3D characteristics, which would make these people get nearly all zero scores when the scoring criteria of MoCA is strictly executed in the CCT, and lead to undermining the validity of visuospatial ability assessment. Therefore, our team introduced a fine scoring standard for CCT. This new criterion classifies drawings based on continuity, presence of two-dimensional (2D) features, and emergence of three-dimensional attributes, establishing a four-level scoring system ( As described in Table 1).Furthermore, the team primarily analyzed static images of cube copying produced by participants, using an edge gradient-based contour extraction algorithm to assign scores. As mentioned before, dynamic trajectories provide richer temporal information than static images. Therefore, the current scoring method for cube copying ability based on static images urgently needs improvement.

In order to solve the above problems, this study proposes a fine assessment method for cube copying ability based on the DH-SCSM-BLA method. The main contributions are as follows:

(1) A fine scoring method was proposed to address the inaccuracy and incompleteness of the original MoCA cube copying test, which relies on a binary classification (0 or 1 scores).

(2) A Spatial-Cosine-Similarity & Motion (SCSM) feature extraction method based on dynamic handwriting segmentation is proposed, encompassing both spatial and motion features of segmented trajectories;

(3) Comparative analyses of cube copying test scores are conducted from three perspectives (MCI patients vs. healthy controls, different age groups, and different educational levels), revealing significant discriminative patterns;

(4) Statistical significance analyses are performed between each cognitive indicator score of Cogni-CareV3.0 App (including the CCT visuospatial cognitive ability score) and two factors (age and educational level). The results indicated that all cognitive metric scores derived from the Cogni-CareV3.0 App are significantly negatively correlated with age, but significantly positively correlated with educational level.

## 2. Method

As shown in Fig. 1, the framework of the proposed method contains three modules: (1) Data modeling, (2) Feature extraction, (3) Classification recognition. The Cogni-CareV3.0 software independently developed by our team was used to collect dynamic handwriting data of CCT. Then, the spatial and motion features of segmented dynamic handwriting were extracted, and feature matrix with unequal dimensions were normalized. Finally, a bidirectional long short-term memory network model combined with attention mechanism (BiLSTM-Attention) was adopted for classification.

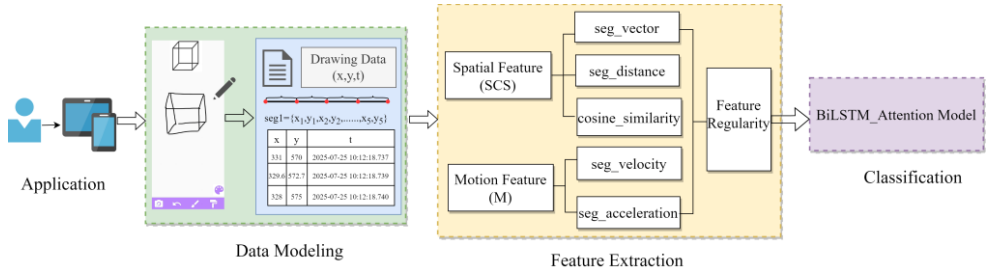

**Fig.1.** Block diagram of the proposed method

## 2.1 Handwriting data mathematical modeling

The dynamic handwriting of the cube copying is characterized by the coordinates (x,y) and time t as core parameters. The original handwriting trajectory dataset is defined as TR, where TR=[$TR_1, TR_2,...,TR_k,...,TR_N$], with k denoting the test sample index and N the total number of test samples. Each sample $TR_k$ consists of $TR_k = [S_{k,1}, S_{k,2}, \ldots, S_{k,i}, \ldots S_{k,L}]$, where $S_{k,i} = (x_{k,i}, y_{k,i}, t_{k,i})$, with $S_{k,i}$ denoting the coordinates and time values at the i-th moment of the k-th sample. The time *t* is measured in milliseconds, and *L* represents the handwriting length (number of handwriting points). Since the number of handwriting points in cubic trajectories varies across individuals, *L* is a variable, resulting in a data dimension of *3L* per sample.

The data acquisition and visualization of single-sample modeling are described as follows: participants complete the CCT using the Cogni-CareV3.0 App, which saves all handwriting information (TR). Fig. 2 denotes the $TR_k$ representation process, where the handwriting coordinates of a specific cube region are shown in the red box. The cube shape is reconstructed from the acquired handwriting data to visually demonstrate the modeling process. Different colored line segments in the figure are generated sequentially based on the timestamp of handwriting points, corresponding to different drawing phases. The red arrows indicate the directions of the first and second strokes.

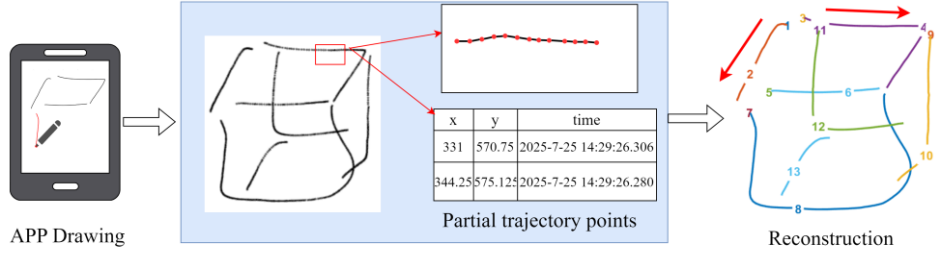

**Fig. 2.** the model of handwriting data

## 2.2 Feature extraction based on SCSM

The original handwriting trajectory data $(x_k, y_k, t_k)$ are processed through a segmentation method to extract 12-dimensional spatial features and 2-dimensional motion features, denoted as SCS features and M features respectively. The SCS features include 10-dimensional segment coordinate vectors, 1-dimensional intra-segment distance, and 1-dimensional cosine similarity. The M features consist of 1-dimensional segment average velocity and 1-dimensional acceleration. These spatial and motion features are combined to form a 14-dimensional feature vector termed SCSM. The feature extraction process is denoted with the k-th sample $TR_k$.

### 2.2.1 Spatial features

The handwriting trajectory coordinates are segmented into fixed-length segments of 5 points, where consecutive segments are connected end-to-start and non-overlapping. Fig. 3 presents a schematic diagram of the handwriting trajectory coordinates. Specifically, Fig. 3(a) and Fig. 3(b) denote the coordinate relationships between segments and within a single segment, respectively.

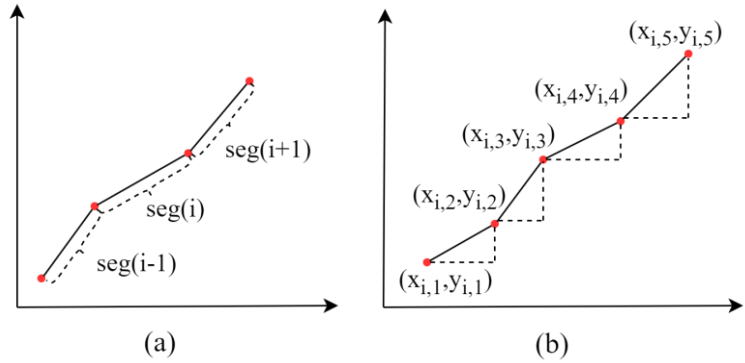

**Fig. 3.** Segment Coordinate diagram. (a) Inter-segment diagram, (b) Intra-segment diagram.

The segment vector (seg_vector) is obtained by converting the intra-segment coordinates into a 10-dimensional vector as shown in Fig.3(b),

$$seg_i = \{x_{i,1}, y_{i,1}, x_{i,2}, y_{i,2}, \dots, x_{i,5}, y_{i,5}\} \quad (1)$$

where i represents the segment index, and the numbers 1, 2,..., 5 represent the indices of the coordinate points within the segment.

The sum of intra-segment distances (seg_distance) is the cumulative Euclidean distance between all adjacent points within a single segment. The calculation formula for the intra-segment distance sum $dis_i$ is:

$$dis_i = \sqrt{\sum_{j=1}^{4}(x_{i,j+1} - x_{i,j})^2 + (y_{i,j+1} - y_{i,j})^2} \qquad (2)$$

where $(x_{i,j}, y_{i,j})$ denotes the coordinates of the j-th point in i-th segment (j=1,2,...,5).

Cosine similarity (cos_similarity) quantifies the directional consistency between trajectory segments by calculating the cosine value of the angle between them. A value closer to 1 indicates that the segments are aligned in the same direction, a value approaching 0 signifies perpendicular alignment, and a value near -1 denotes opposite directions. This method typically adopts vector space similarity metrics[38], and the inter-segment vector diagram is denoted in Fig. 4.

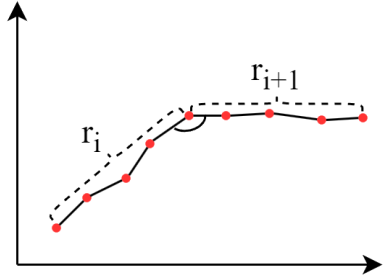

**Fig. 4.** Inter-segment Angle diagram

Given two adjacent segment vectors $r_i = \{x_{i,1}, y_{i,1}, \ldots\ldots, x_{i,5}, y_{i,5}\}$ and $r_{i+1} = \{x_{i+1,1}, y_{i+1,1}, \ldots\ldots, x_{i+1,5}, y_{i+1,5}\}$, the inter-segment cosine similarity feature vector $sim$ can be obtained as follows:

$$sim = \frac{r_i \cdot r_{i+1}}{\|r_i\|\|r_{i+1}\|} = \frac{\sum_{j=1}^{5}(x_{i,j} \cdot x_{i+1,j} + y_{i,j} \cdot y_{i+1,j})}{\sqrt{\sum_{j=1}^{5}(x_{i,j}^2 + y_{i,j}^2)} \cdot \sqrt{\sum_{j=1}^{5}(x_{i+1,j}^2 + y_{i+1,j}^2)}} \qquad (3)$$

where $r_i \cdot r_{i+1}$ denotes the dot product of two segment vectors, while $\|r_i\|$ and $\|r_{i+1}\|$ represent their respective magnitudes.

### 2.2.2 Motion features

Motion features include the average velocity and acceleration. Segment velocity can measure the speed of handwriting point movement, and the average segment velocity $v_i$ can be obtained using the intra-segment distance sum $dis_i$ calculated by Equation (2):

$$v_i = \frac{dis_i}{\Delta t_i} \qquad (4)$$

where the intra-segment time difference $\Delta t_i = t_{i,end} - t_{i,start}$, where $t_{i,end}$ represents the time of the last point in the current $i\text{-}th$ segment, and $t_{i,start}$ denotes the time of the first point in the same segment.

Segment acceleration can capture sudden changes in handwriting velocity. It is derived by calculating the velocities of adjacent points within the segment and their corresponding time intervals, with the expression for segment acceleration $a_i$ given by:

$$a_i = \frac{v_{i,j+1} - v_{i,j-1}}{t_{i,j+1} - t_{i,j-1}} \quad (5)$$

where $v_{i,j+1}$ and $v_{i,j-1}$ represent the instantaneous velocities of the $(j+1)$ and $(j-1)$ points in the $i$-th segment of the k-th sample, while $t_{i,j+1}$ and $t_{i,j-1}$ denote the timestamps of the coordinate points $(j=2,3,4)$.

### 2.2.3 Feature Normalization

The five aforementioned features are concatenated into a composite feature $f$ according to Equation (6). However, this concatenation method has a limitation: since the original handwriting sequence L varies in length across samples, the number of segments obtained after segmentation also differs per sample. Meanwhile, the feature dimension extracted from each segment remains fixed—resulting in composite features of unequal lengths and thus forming feature vectors with non-uniform dimensions. To address this issue, this paper achieves feature dimension unification through standardized length intervals and resampling. The specific operations are as follows: A fixed-length interval of $[14, L_{std}]$ is established for the standardized handwriting sequence. Here,14 is the intrinsic dimension of the feature, and $L_{std}$ is the standardized sequence length, dictated by the mean length of the sample sequences. Resampling is then performed on the original sequence L：if L<$L_{std}$ the sequence is extended via interpolation; if L>$L_{std}$ it is compressed through downsampling. After these processes, the feature set of all samples is expressed as shown in Equation (7):

$$f = [seg, dis, sim, v, a]_{14 \times L_{std}} \quad (6)$$

$$F = \begin{bmatrix} f_1 \\ f_2 \\ \vdots \\ f_N \end{bmatrix} \quad (7)$$

where $f_1, f_2, ..., f_N$ represent the extracted features of the 1st, 2nd,..., and Nth samples after normalization, and each $f_d$ ($d=1,2,…,N$) is a feature matrix with dimension of $14 \times L_{std}$.

### 2.3 Classification based on BiLSTM-Attention

For the extracted feature vectors, this study adopts a bidirectional long short-term memory network model (BiLSTM-Attention) for classification. Initially proposed by Zhou et al.[39] for natural language processing tasks, this model integrates bidirectional long short-term memory (BiLSTM) with an attention mechanism, enabling effective capture of critical information from sequential features. This paper combines the previously proposed features to obtain 3 combined features, namely SCS, M, and SCSM features. SCS include segment coordinates, segment distance, and inter-segment cosine similarity; M combines segment velocity and acceleration; and SCSM merges the first two types of features. Each combined feature is input into the BiLSTM-Attention model for classification, with the optimal feature combination identified through evaluation of classification accuracy.

The network model designed in this paper (Fig. 5) comprises an input layer, a bidirectional LSTM layer, an attention layer, and an output layer. The specific workflow is as follows: The input layer receives a standardized feature sequence $F=[F_1,F_2,...,F_t,...F_n]$ (where n denotes the dimension of sample features) and feeds it into a two-layer LSTM. The bidirectional LSTM layer processes the original feature sequence through a forward layer and the reversed feature sequence through a backward layer. Outputs from the two layers are concatenated to generate a hidden layer state matrix $H=[h_1,h_2,...,h_t,...,h_n]$ (where $h_t$ represents the bidirectional hidden state at time step t, $i$ denotes the hidden layer dimension of the LSTM neurons set to 128, and $C_t$ and $h_t$ respectively denote the cell state and hidden state of the LSTM at time step t). This matrix serves as the input to the attention layer, whose core mechanism computes the weight relationships between "query (Q), key (K), and value (V)" [40]. The layer first performs linear transformations on $H$ to obtain Q, K, and V. Subsequently, the SoftMax layer calculates the attention weight for each time step using Scaled Dot-Product Attention, followed by a linear mapping to generate the intermediate output. The output layer employs regularization to mitigate overfitting, and the features are mapped to the category space via a fully connected layer to produce the final classification result.

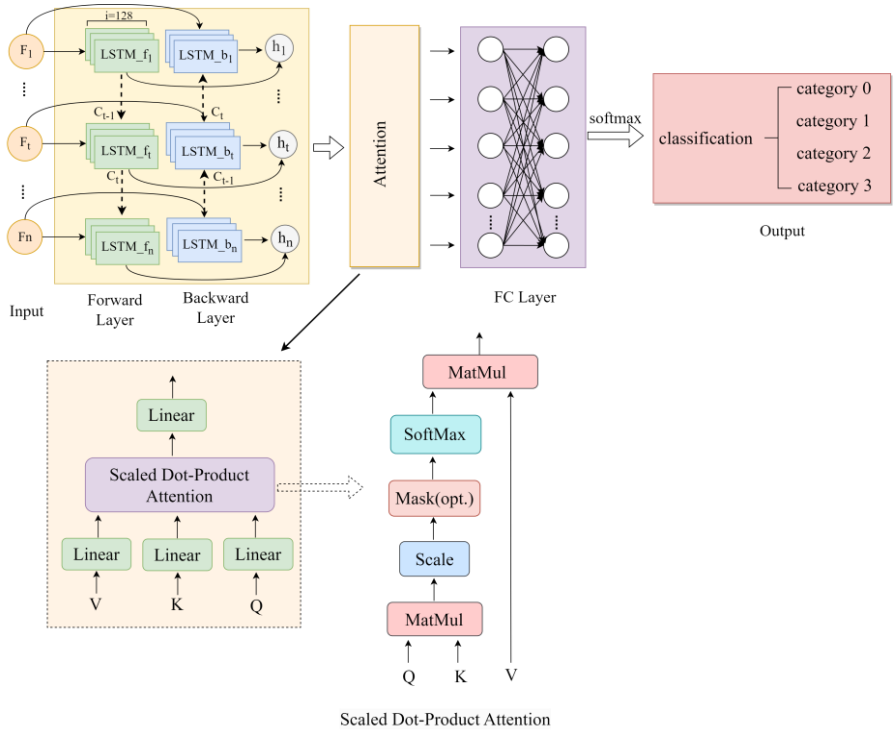

**Fig. 5.** BiLSTM-Attention network structure

## 3. Experiment and result

This study adheres to the principles of academic ethics: the personal information of all participants is strictly confidential, and data collection was conducted with the informed consent of the subjects.

The Cogni-CareV3.0 App collects comprehensive test data through its ten task modules, namely short-term memory, executive function, naming animals, visuospatial cognition, calculation, attention, language, abstract, delayed memory and orientation, where the visuospatial module mainly includes the CCT and clock recognition. Data were collected from participants recruited from a nursing home, a village, and a university. All subjects completed the full set of game-based tests using the Cogni-CareV3.0, forming a self-collected dataset comprising 224 participants. The gender ratio, age distribution, and educational background of the participants are presented in Fig. 6(a), 6(b), and 6(c), respectively, while the distribution of participants MCI and Healthy Control is denoted in Fig. 6(d).

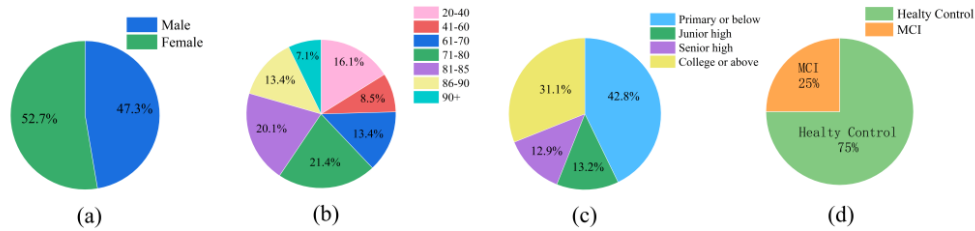

**Fig .6.** Distribution of participants' information.(a) Distribution of Sex; (b) Distribution of Age; (c) Distribution of Educational Level; (d) Distribution of MCI and Healthy Control.

## 3.1 Acquisition of handwriting data

As previously described, the study utilized handwriting data from the visuospatial task of the selected dataset to evaluate the potential of dynamic handwriting features in AD-related visuospatial cognition. The CCT in the Cogni-CareV3.0 App is denoted in Fig. 7, where participants draw standard cubes on the game interface, with their handwriting data being stored in real-time in the database.

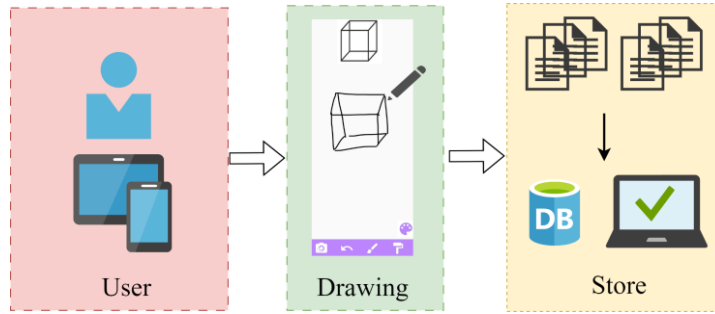

**Fig.7.** Block diagram of handwriting data acquisition

## 3.2 Fine scoring criteria for CCT

As mentioned earlier, the elder with lower educational level generally score 0 point based on the scoring criteria for visuospatial tasks in the MOCA. Therefore, in this study, sample labels are defined based on the accuracy and structural integrity of the drawings completed by participants during the CCT, and the data were categorized into four levels: Category 0 (0 score), Category 1 (1 score), Category 2 (2 score), and Category 3 (3 score). The specific classification rules are detailed in Table 1.

**Table 1.** Classification Rules

| Category | Description |
|---|---|
| 0 score | Unable to form a two-dimensional structure, only partial lines, scattered line segments, or a circle, a randomly smeared black blob, etc. can be drawn. |
| 1 score | The drawn figures have 2D structural characteristics and no 3D stereoscopic structure, such as only one or more unconnected rectangles being drawn. |
| 2 score | Able to draw a cube with multiple surfaces; relatively complete shape, basic 3D cube structure, slight proportional imbalance, or 1-2 missing edge lines. |
| 3 score | Requires 12 edge lines fully drawn, correct vertex connections, complete structure, clear lines, distinct cubic spatial structure, and no obvious deformation. |

All samples in the current dataset were classified according to the four-level classification criteria in Table 1. The training set, validation set, and test set were divided in a 8:1:1 ratio (Table 2).

**Table 2.** Number of Sample Categories

| Data | Total | Category 0 | Category 1 | Category 2 | Category 3 |
|---|---|---|---|---|---|
| Total set | 224 | 48 | 67 | 67 | 42 |
| Training set | 178 | 38 | 53 | 53 | 34 |
| Validate set | 23 | 5 | 7 | 7 | 4 |
| Test set | 23 | 5 | 7 | 7 | 4 |

## 3.3 Results analysis

The experiments were conducted on a server equipped with NVIDIA RTX A4000 GPU and Intel processors (Table 3), with the operating environment based on the PyTorch 2.0.0 + CUDA 11.8 framework.

**Table 3**. Environment parameter

| | Component | Version |
|---|---|---|
| Hardware | GPU | NVIDIA RTX A4000 GPU(16GB) |
| | CPU | Intel Xeon(R) W-2223 CPU |
| | GPU memory | 32GB |
| Software | OS | Windows 10 |
| | Python | Python3.9 |
| | Pytorch | 2.0.0 |

Under the aforementioned experimental conditions, the training set comprised 178 dynamic handwriting samples, where the input feature is defined as $F$. Given the small scale of the dataset, a 5-fold cross-validation method was employed to evaluate the model's generalization performance. This study evaluates the performance of the classification method using three metrics: Accuracy, Precision, and F1 score. As presented in Table 4, experimental results indicate that Experiment 3 achieved the optimal performance. The model reached a stable performance state after 30 epochs. Notably, the maximum performance is achieved with the optimal configuration, where the number of LSTM layers, hidden layer dimension, attention layer input dimension, and learning rate are set to 2, 128, 256, and 0.005, respectively.

Table 4. BiLSTM-Attention Model Parameters

| Number | LSTM Layer | Hide_dim | Attention_dim | Lr | Epoch | Accuracy/% | Expense/s |
|---|---|---|---|---|---|---|---|
| 1 | 2 | 64 | 128 | 0.005 | 100 | 78.26 | 2.48 |
| 2 | 2 | 128 | 256 | 0.001 | 100 | 82.61 | 2.80 |
| 3 | 2 | 128 | 256 | 0.005 | 100 | 86.96 | 2.88 |
| 4 | 2 | 128 | 256 | 0.01 | 100 | 82.61 | 3.09 |
| 5 | 1 | 128 | 256 | 0.005 | 100 | 78.26 | 2.71 |

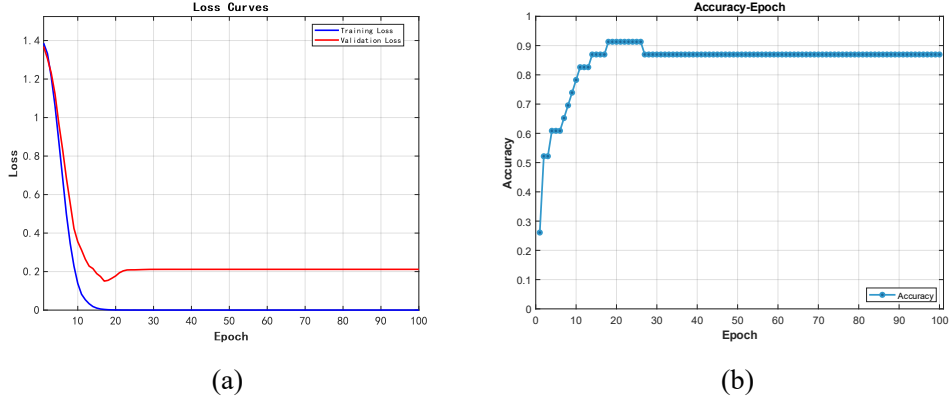

**Fig. 8.** Loss and Accuracy Curves of BiLSTM-Attention Model. (a) Loss Curves , (b) Accuracy Curves.

To avoid limitations of single classification methods and comprehensively evaluate model performance, this study compares the recognition performance of multiple classifiers on the feature SCSM, including GRU, BiGRU, and Transformer models. Table 5 presents the classification performance of the feature across different models, with core metrics such as accuracy, precision, and F1 score used for evaluation.

Table 5. Accuracy of different classification algorithms

| Model | Accuracy(%) | Precision(%) | F1 score |
|---|---|---|---|
| RF | 69.57 | 70.79 | 0.692 |
| DT | 60.29 | 60.55 | 0.602 |
| KNN | 47.83 | 51.45 | 0.452 |
| DNN | 83.33 | 84.49 | 0.822 |
| Transformer | 73.91 | 73.91 | 0.737 |
| BiGRU | 82.61 | 84.96 | 0.822 |
| GRU | 69.57 | 71.62 | 0.698 |
| LSTM | 73.91 | 75.36 | 0.738 |
| LSTM-Attention | 78.26 | 78.41 | 0.780 |
| BiLSTM | 73.91 | 75.54 | 0.743 |
| BiLSTM-Attention | 86.96 | 88.37 | 0.866 |

Given the limited sample size of the dataset, we initially selected GRU and BiGRU models—both designed for small sample learning—as baseline benchmarks. However, the experimental results revealed that the BiLSTM-Attention model achieved the highest classification accuracy.

The three obtained combined features, namely SCS, M, and SCSM were input into the BiLSTM-Attention model for ablation experiments. The classification results of these features on the dataset (as shown in Table 6) demonstrate that the SCSM combination feature achieves higher recognition accuracy than individual features.

Table 6. Comparison of feature ablation experiments

| Feature Set | Dimension | Accuracy(%) | Precision(%) | F1 score |
|---|---|---|---|---|
| SCS | 12 | 63.24 | 64.05 | 0.634 |
| M | 2 | 47.83 | 44.93 | 0.438 |
| SCSM | 14 | 86.96 | 88.37 | 0.866 |

## 4. Discussion

This study demonstrates the advantages of segmented feature extraction methods in evaluating visuospatial cognitive ability through dynamic handwriting analysis by comparing feature combinations with models. Building on this foundation, this chapter explores three key aspects: validating the effectiveness of segmented feature extraction, examining the distribution of visuospatial cognitive scores across different population groups, and conducting significance analysis of age and educational level effects on visuospatial cognitive performance.

### 4.1 Comparison between DH-SCSM-BLA method and similar literature

To comprehensively evaluate the effectiveness of the proposed segmented feature extraction method, we selected methods from reference [26] and [27] as baseline benchmarks. Both studies focused on analyzing writing behaviors in neurological disorders, employing kinematic features for modeling. Notably, reference [27] further integrated geometric and nonlinear dynamics approaches to construct feature models, achieving classification performance for both PD patients and healthy individuals.

Since the extraction of geometric features (such as altitude_diff, azimuth_diff, and other parameters) usually relies on specific devices like digital tablets, while the data conditions of this study cannot directly obtain such features, to ensure the fairness and feasibility of the comparison, only directly extractable kinematic features (including curvature, writing speed, average acceleration) and nonlinear dynamic features (including sample entropy, Hurst index, correlation dimension) were selected for comparison. The experimental results are shown in Table 8.

Table 7. Feature Method Description

| Feature | Name | Description |
|---|---|---|
| Kinem | curvature | The curvature of the local path |
| | speed | Changes in motion speed |
| | total_time | Total time of the path |
| | total_distance | Total distance |
| | Basic Statistical Features | Statistical characteristics of speed and acceleration, such as standard deviation, maximum and minimum values, and mean |
| | Peak speed | Indicates speed fluctuation |
| | Acceleration peak | Indicates acceleration fluctuation |
| NLD | Sample entropy | Trace the irregularity of the sequence |
| | Hurst | Track handwriting sequence persistence or anti-persistence |
| | Correlation Dimension | The complexity of system dynamics |

Table 8. Experimental Results

|  | Model | Accuracy(%) | Precision(%) | F1 score |
|---|---|---|---|---|
| Method of [27] | BiLSTM-Attention | 65.22 | 65.60 | 0.625 |
|  | RF | 56.52 | 57.25 | 0.523 |
| Method of [26] | BiLSTM-Attention | 52.17 | 55.60 | 0.502 |
|  | RF | 60.87 | 63.77 | 0.592 |
| DH-SCSM-BLA | BiLSTM-Attention | 86.96 | 88.37 | 0.866 |
|  | RF | 69.57 | 70.79 | 0.692 |

As shown in Table 8, our proposed method was compared with the feature extraction approaches from reference [26] and reference [27] on the same dataset. To eliminate the interference of different classifiers, ablation experiments were conducted using the same model for both training and testing. The results demonstrate that our method outperforms the baselines across all evaluation metrics, which highlights the superior data representation capability of our SCSM feature extraction approach. For the relatively simple features proposed in reference [26], the RF model achieved better performance, as the complex BiLSTM-Attention model may suffer from performance degradation due to overfitting.

Valla et al.[26] extract kinematic and spatiotemporal features of handwriting trajectories but fail to capture local motion information. While the method proposed by Rios-Urrego et al.[27] captures the basic physical attributes of writing movements and the irregularities of time series, it loses the original spatial position (x, y) information of the sequence data and does not consider the trajectory trends within the time period. In contrast, the SCSM feature extraction method proposed in this paper focuses on the spatial morphological information of (x,y,t) sequences. Specifically, the SCSM method introduces a "segmented trajectory" strategy to quantify the continuity of trajectory motion.

## 4.2 Distribution of CCT Scores Among Different Groups

To further examine the distribution characteristics of visuospatial cognitive scores across populations, we analyzed the module's distribution from three perspectives—MCI versus Healthy Control, age, and educational level—based on cognitive test scores (0-3 scores), as shown in Fig. 9.

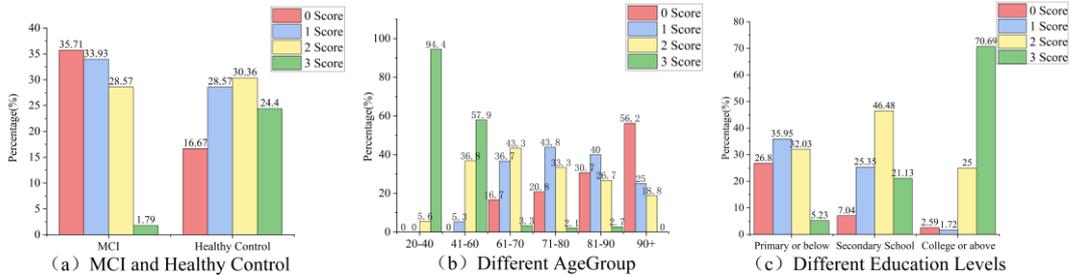

**Fig. 9.** Distribution of CCT scores across different groups.(a) Distribution of MCI and Healthy Control. (b) Distribution of Different Age Group. (c) Distribution of Different Educational Levels.

As shown in Fig. 9(a), the Healthy Control group exhibits a relatively balanced and high-scoring distribution in the visuospatial task, while the MCI group shows a distinct low-scoring tendency—with 69.64% of participants scoring 0 or 1 score. This indicates that the cube-drawing visuospatial task is particularly sensitive to MCI, as individuals with MCI exhibit greater difficulties in this task compared to cognitively normal counterparts. Fig. 9(b) presents age-specific scores distribution patterns: task scores begin to decline in the 41-60 age group, with the rate of 3-scores dropping to 3.3% in the 61–70 age group. Collectively, these findings confirm a consistent downward trend in task performance across successive age groups. Fig. 9(c) highlights notable differences in task performance by educational level: individuals with lower educational level predominantly achieved low scores, reflecting limited cube-drawing proficiency, while middle school graduates exhibited relatively consistent performance. Notably, over 70% of university-educated participants successfully completed the task, suggesting that higher education may significantly enhance spatial visualization and expression skills—potentially through coursework in geometry, cartography, and engineering-related disciplines.

## 4.3 Significance Analysis On Scores of Each Cognitive Task

This study conducted significance analysis on scores of each cognitive task to age and educational level respectively. The Pearson correlation coefficient was calculated using the covariance matrix of data to evaluate the relationship strength between the two variables[41]. Based on the population distribution characteristics, the educational level was quantified as years of education for Pearson correlation coefficient.

As shown in Tables 9 and 10, there is a significant negative correlation between scores of each cognitive task and participants' age, where elder participants demonstrated lower cognitive scores. Conversely, scores of each cognitive task except Delayed Recall task showed a positive correlation with levels of education.

**Table 9.** Significance Analysis On Scores of Each Cognitive Task and Age

| Task | Name | r-value | P-value | Significance |
|---|---|---|---|---|
| Task1 | Short-term Memory | -0.523 | <0.001 | *** |
| Task2 | Trail Making | -0.608 | <0.001 | *** |
| Task3 | Naming animals | -0.361 | <0.001 | *** |
| Task4 | Clock | -0.405 | <0.001 | *** |
| Task5 | Cube Copy Test | -0.708 | <0.001 | *** |
| Task6 | Calculation | -0.573 | <0.001 | *** |
| Task7 | Attention | -0.633 | <0.001 | *** |
| Task8 | Language | -0.426 | <0.001 | *** |
| Task9 | Abstract | -0.555 | <0.001 | *** |
| Task10 | Delayed Recall | -0.486 | <0.001 | *** |
| Task11 | Orientation | -0.522 | <0.001 | *** |

Note: ***: p<0.001 (extremely significant), **: p<0.01 (highly significant), *: p<0.05 (significant), "-" indicates no statistical significance; |r|≥0.7 indicates strong correlation, 0.3≤|r|<0.7 indicates moderate correlation, and |r|<0.3 indicates weak correlation.

Table 10. Significance Analysis On Scores of Each Cognitive Task and Educational level

| Task | Name | r-value | P-value | Significance |
|---|---|---|---|---|
| Task1 | Short-term Memory | 0.358 | <0.001 | *** |
| Task2 | Trail Making | 0.398 | <0.001 | *** |
| Task3 | Naming animals | 0.168 | 0.0122 | * |
| Task4 | Clock | 0.344 | <0.001 | *** |
| Task5 | Cube Copy Test | 0.537 | <0.001 | *** |
| Task6 | Calculation | 0.343 | <0.001 | *** |
| Task7 | Attention | 0.286 | <0.001 | *** |
| Task8 | Language | 0.405 | <0.001 | *** |
| Task9 | Abstract | 0.307 | <0.001 | *** |
| Task10 | Delayed Recall | 0.109 | 0.1031 | - |
| Task11 | Orientation | 0.372 | <0.001 | *** |

Note: Same as Table 9.

Furthermore, when performing significance analyses on cognitive task scores across different age groups and educational levels, the results of the significance analysis on scores of the cognitive task were found to be highest in the CCT.

## 5. Conclusion

In this study, we used a self-developed Cogni-CareV3.0 App to collect handwriting data and organized it into a dynamic handwriting dataset. To effectively preserve the spatiotemporal characteristics of original trajectories, we propose a fine assessment method which named DH-SCSM-BLA based on dynamic trajectories. This method constructs segmental feature combinations of trajectories using a "segmentation strategy" and normalizes them into equal-dimensional feature matrices, which are then input into a BiLSTM-Attention model. Experimental results demonstrate that the method achieved recognition rates of 86.69%, exhibiting superior classification performance compared to single-feature extraction methods. Collectively, this study provides a novel approach for the early screening of MCI.

However, this study has certain limitations. As discussed earlier, the CCT may be influenced by participants' educational level: specifically, participants with lower years of education tend to receive 0 score, while those with higher education may generally obtain higher scores due to their prior training and experience related to geometric figures. As reported by Galaz et al.[42], the pentagon copying test can be used to quantify cognitive decline, which may mitigate the impact of 3D geometric figures on participants. In future research, supplementary assessment tools such as the pentagon copying test could be incorporated to conduct a comprehensive analysis of participants' visuospatial cognitive abilities, thereby reducing the interference of educational background as a confounding variable on the evaluation results.

## 6. Acknowledgments

This study was financially supported by: (1) Key project of Natural Science funded by Education Department of Anhui Province of 2019 (KJ2019ZD56), China. (2) Clinical New Technology Incubation Project of Changhai Hospital (2024XC108), China. We thank iFlytek for its technical support and free resources, and also thank the elderly residents at a nursing home and in a rural community for their strong support and assistance, as well as the nursing home staff for their significant coordination, which greatly facilitated the smooth progress of the research.